\documentclass[conference]{IEEEtran}
\IEEEoverridecommandlockouts

\usepackage{amsmath,amssymb,amsfonts}
\usepackage{algorithmic}
\usepackage{graphicx}
\usepackage{textcomp}
\usepackage[dvipsnames]{xcolor}
\def\BibTeX{{\rm B\kern-.05em{\sc i\kern-.025em b}\kern-.08em
    T\kern-.1667em\lower.7ex\hbox{E}\kern-.125emX}}

\usepackage[pagebackref=true,breaklinks=true,colorlinks,bookmarks=false]{hyperref}
\usepackage[numbers,sort&compress]{natbib}
\usepackage{booktabs}
\usepackage{colortbl}
\usepackage{bbm}
\definecolor{myorange1}{RGB}{255, 0, 0}
\definecolor{mygray}{gray}{.9}
\usepackage{pifont}
\usepackage{multirow}
\usepackage{makecell}
\usepackage{array}
    
\begin{document}

\title{Spatial-Temporal Perception with Causal Inference for Naturalistic Driving Action Recognition}



\author{
  Qing Chang\textsuperscript{1*}, 
    Wei Dai\textsuperscript{2*}, 
    Zhihao Shuai\textsuperscript{3}, 
    Limin Yu\textsuperscript{2}, 
    Yutao Yue\textsuperscript{3†}%
    \thanks{*These authors contributed equally.}%
    \thanks{†Corresponding author: yutaoyue@hkust-gz.edu.cn}%
    \vspace{0.4cm}\\
    \textsuperscript{1}School of Mechanical Engineering, Nanjing University of Science and Technology
    \\\textsuperscript{2}School of Advanced Technology, Xi'an Jiaotong-Liverpool University
    \\\textsuperscript{3}The Hong Kong University of Science and Technology (Guangzhou)%
}

\maketitle
\begin{abstract}
    Naturalistic driving action recognition is essential for vehicle cabin monitoring systems. However, the complexity of real-world backgrounds presents significant challenges for this task, and previous approaches have struggled with practical implementation due to their limited ability to observe subtle behavioral differences and effectively learn inter-frame features from video.
    In this paper, we propose a novel Spatial-Temporal Perception (STP) architecture that emphasizes both temporal information and spatial relationships between key objects, incorporating a causal decoder to perform behavior recognition and temporal action localization. Without requiring multimodal input, STP directly extracts temporal and spatial distance features from RGB video clips. Subsequently, these dual features are jointly encoded by maximizing the expected likelihood across all possible permutations of the factorization order. By integrating temporal and spatial features at different scales, STP can perceive subtle behavioral changes in challenging scenarios. Additionally, we introduce a causal-aware module to explore relationships between video frame features, significantly enhancing detection efficiency and performance. We validate the effectiveness of our approach using two publicly available driver distraction detection benchmarks. The results demonstrate that our framework achieves state-of-the-art performance.
\end{abstract}


\section{Introduction}
Naturalistic driving action recognition (DAR) is a critical component of vehicle cabin monitoring systems. 
Its primary objectives are distracted behavior detection and temporal action localization (TAL) within untrimmed video sequences, both of which are vital for improving driving safety and fostering effective driver-vehicle interaction.

\begin{figure}[!t]
    \centering
    \includegraphics[width=.9\linewidth]{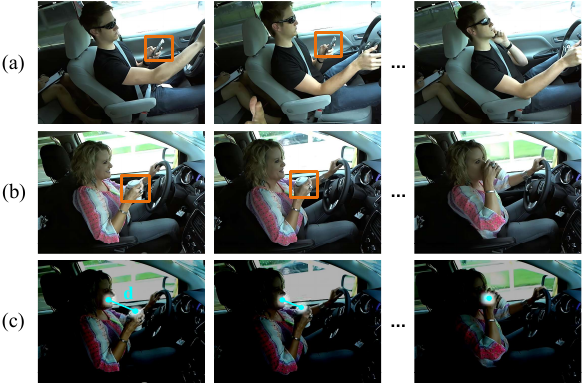}
    \caption{Illustrations of challenging cases in driving action recognition.
    (a) Calling and (b) drinking scenes, where the objects are partially visible and the lighting conditions are unstable. (c) Variations in the distance $d$ between key points assist in identifying behavior categories and temporal localization.
    }
    \vspace{-0.4cm}
    \label{fig:intro}
\end{figure}

Recent advancements in DAR have been propelled by the powerful representation capabilities of deep learning~\cite{adewopo2023review}. Several approaches build on general human action recognition backbones, leveraging 3D convolutional neural networks (CNNs)~\cite{chen2023skateboardai} and vision transformers~\cite{wang2023videomae}. Among these, the temporal shift module (TSM)~\cite{lin2019tsm} has demonstrated effectiveness in learning features from adjacent frames.

Despite this progress, DAR remains a challenging task due to the complex nature of the vehicle cabin environment, which provides limited distinguishing features.
As shown in Fig.~\ref{fig:intro}, drivers often exhibit highly similar movements of body parts (\emph{e.g.}, eating and drinking), which can easily confuse detectors. Additionally, variable lighting conditions in the cabin and the duration of input video clips pose further challenges in modeling long-sequence feature relationships.

\begin{figure*}[!t]
\centering
\includegraphics[width=0.9\linewidth]{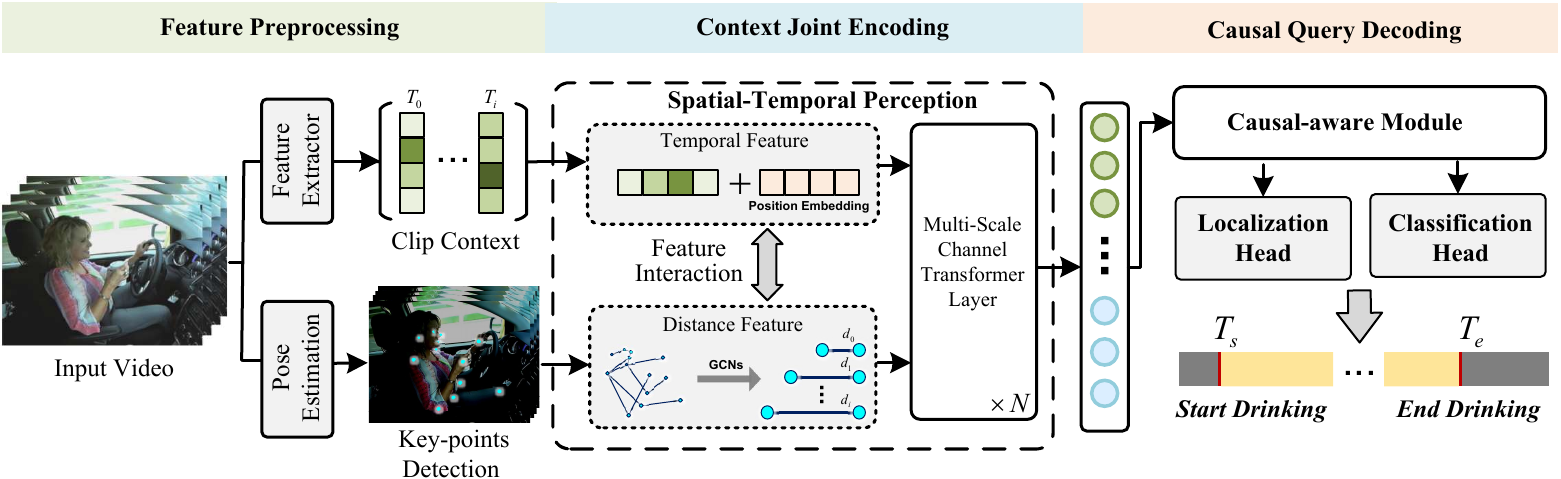}
\caption{The overall architecture of STP.}
\vspace{-0.3cm}
\label{fig:pipe}
\end{figure*}

Several studies have sought to address these challenges.  Khan et al.~\cite{khan2022supervised} utilized depth and infrared inputs using a late fusion method to improve the robustness of driving behavior detection. Ma et al.~\cite{ma2022real} proposed a multi-scale attention module for multi-view image fusion, while Jiang et al.~\cite{kuang2023mifi} developed a multi-camera DAR model that trains a single-camera feature extractor to boost performance. However, these methods are limited by the requirement for additional input types and depend on single-mode classification pipelines, which reduce their practicality in real-world hardware environments and compromise efficiency. Additionally, they often neglect the temporal correlations between frames.

After thorough analysis, we argue that two critical features from video merit attention: temporal information and the spatial distance between interest objects. Temporal information is directly derived from video clips, providing fine-grained visual features essential for action recognition~\cite{he2023reti,guan2024asy}. Furthermore, as illustrated in Fig.~\ref{fig:intro}(c), changes in the distance between a driver’s mouth and a cup offer cues for identifying the start and end of actions. By aligning and fusing these two feature types, the model can better focus on key regions to perceive the action, thereby mitigating irrelevant interference. 

To this end, we present a novel Spatial-Temporal Perception (STP) network that emphasizes both temporal information and spatial relationships between interest objects, incorporating a causal decoder to perform behavior recognition and temporal action localization. Without relying on multi-view and multimodal input, STP directly extracts temporal and spatial distance features from RGB video clips. These dual features are jointly encoded by maximizing the expected likelihood across all possible permutations of the factorization order. By integrating temporal and spatial features, STP is capable of detecting subtle behavioral changes in challenging scenarios. Furthermore, we introduce a causal-aware module to analyze relationships between video frame features, significantly improving detection efficiency and performance.
We validated the effectiveness of our approach using two publicly available driver distraction detection benchmarks:  Drive\&Act and SynDD2. The results demonstrate that our framework achieves state-of-the-art performance in both driver action recognition and temporal action localization tasks.

\section{Methodology}\label{sec:method}
\subsection{Overview}\label{sec:overview}
The overall architecture of STP is illustrated in Fig.~\ref{fig:pipe}. Given a video with $T$ frames, denoted as $X \in \mathbb{R}^{T \times  3 \times H \times W}$, the STP aims to integrate the temporal features of video clips with the spatial relationships between key points to enhance driver action recognition and temporal localization. The video clip is first processed in parallel by two heads to extract the clip context and key point positions. These two outputs are then combined, where the clip context is aggregated with position embeddings to generate temporal features, and key points are refined into distance features for each frame using graph convolutional networks (GCNs)~\cite{kipf2016semi}.  
These two types of features are interactively fused to align in space and passed to a stacked multi-scale channel transformer for context encoding. The proposed causal-aware module further explores the relationships between feature sequences. Finally, these hybrid features are decoded by the classification and localization heads to accurately identify the behavior pattern and the start and end frames of action.


\subsection{Context Joint Encoding}\label{sec:context encoding}
In the feature extraction stage, two lightweight extraction networks are employed to obtain dual features. However, the dense temporal features are not aligned with the sparse distance features between nodes extracted by GCNs. To address this misalignment, we first introduce an interaction mechanism to spatially align the dual feature sets. These aligned features are then fed into stacked multi-scale channel transformer layers for fusion and calibration, enabling the integration of both global and local information~\cite{chang2024hybrid,lai2024drive}.

\noindent{\textbf{Dual Feature Interaction.}} 
Feature interaction transfers and aligns features between different modalities. 
Let the $T$-frame dual feature clips be represented as $X^p=\left\{x_t^p\right\}_{t=1}^T$ and $X^d=\left\{x_t^d\right\}_{t=1}^T$, where $x_t^p$, $x_t^d \in \mathbb{R}^{ C\times L} $ denote the temporal and distance features at timestamp $t$, respectively.
The dual features are updated by shifting the last $k$ feature channels of each modality as follows:
\begin{align}
\hat{x}_t^p &= \operatorname{MLP}\left(x_t^p[:-k], x_t^d[-k:]\right), \\
\hat{x}_t^d &= \operatorname{MLP}\left(x_t^d[:-k], x_t^p[-k:]\right),
\end{align}
where $[\cdot,\cdot]$ denotes channel-wise concatenation, and $\operatorname{MLP}$ refers to a fully connected layer.
This process incurs minimal computational cost. As a result, dual-feature interaction efficiently aligns and integrates information across modalities.

\noindent{\textbf{Multi-Scale Encoder.}} 
We adopt a stacked multi-scale channel transformer layer within our recurrence mechanism to facilitate the reuse of hidden states from preceding segments. For a longer sequence $F$ obtained via dual-feature interaction, consider extracting two segments $\tilde{S}=F_{1: t}$ and $S=F_{t+1: 2 t}$ for illustration. We process the initial segment $\tilde{S}$ and retain the resultant content representations $\tilde{H}(m)$ for each layer $m$.
Let $Q=H{(m-1)}$ and $K,V = \left[\tilde{H}{(m-1)}, H{(m-1)}\right]$. When processing the subsequent segment, the attention update, integrating memory, can be formulated as follows:
\begin{equation} \label{}
H^{(m)} = \operatorname{Softmax} \left(\frac{QK^T}{\sqrt{D} } \right)V,
\end{equation}
where $D$ denotes the embedding dimension. Consequently, once the representations $\tilde{H}{(m-1)}$ have been obtained, the attention update operates independently of the variable $\tilde{S}$.

\subsection{Causal Query Decoding}\label{sec:decoding}
\begin{figure}[tb]
    \centering
    \includegraphics[width=0.95\linewidth]{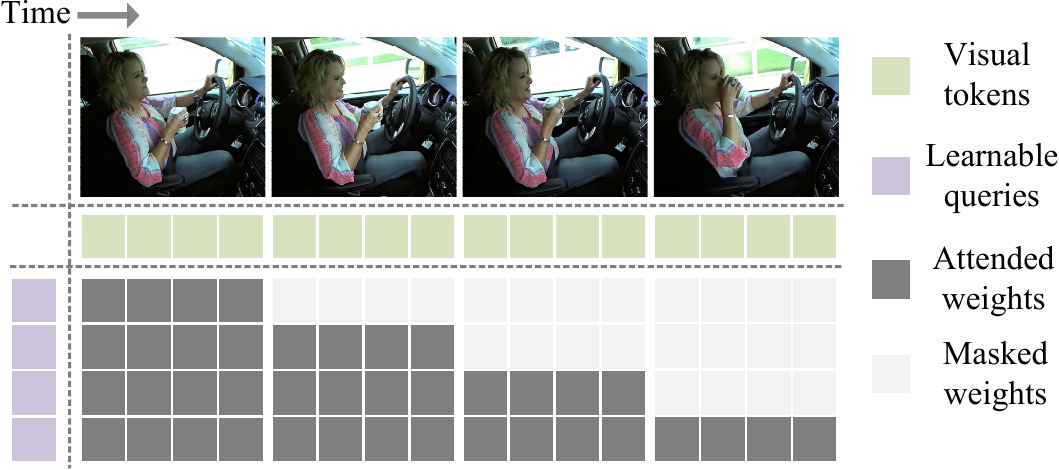}
    \caption{Causal-aware module incrementally exposes video frames to learnable queries to decouple spatial and temporal features.
    }
    \vspace{-0.3cm}
    \label{fig:causal_model}
\end{figure}

\noindent{\textbf{Causal-aware Module.}} 
To ensure the query attends equally to the embeddings of each time frame and fully explores the causal relationships between video frames, we propose a causal-aware module based on cross-attention masks.
Specifically, the output of the module for video embeddings is computed as follows: 
\begin{equation} \label{}
y_i=\frac{\sum_j M_{i j} \exp \left(Q_i K^T\left(x_j\right)\right) V\left(x_j\right)}{\sum_j M_{i j} \exp \left(Q_i K^T\left(x_j\right)\right) \mathbf{1}_L},
\end{equation}
where $M_{ij}$ denotes the mask for the $i$-th query $Q_i$ of the $j$-th frame $x_j$. As current time step $t$ is typically smaller than the number of tokens $n$, $ M_{ij} = 1$ if $i \geq j\left\lfloor\frac{n}{t}\rfloor\right.$, otherwise $M_{ij} = 0$. The $\mathbf{1}_L=[1, \cdots, 1]^T \in \mathbb{R}^{L \times 1}$ is a vector of ones. We illustrate the masking process of our module in Fig.~\ref{fig:causal_model}. This approach ensures that initial queries focus on early visual embeddings, while final queries can access embeddings from various time frames to capture causal relationships across time.

\noindent{\textbf{Prediction Head.}} 
The prediction head consists of a classification head and a localization head. A feed-forward network (FFN)~\cite{vaswani2017attention} is used to predict parameters, and the prediction head is formulated as:
\begin{align}
[\boldsymbol{T_{s}}, \boldsymbol{T_{e}}]=\operatorname{FFN}_{reg }(\mathbf{Q}), \\
\mathcal{C}=\operatorname{FFN}_{cls}(\mathbf{Q}),
\end{align}
where $\boldsymbol{T_{s}}, \boldsymbol{T_{e}}$ represent the start and end frames of the action, and $\mathcal{C}$ denotes the predicted action category.

\noindent{\textbf{Total Loss.}} 
Given matched ground truth labels for the prediction queries, we calculate the corresponding loss for each matched pair. The overall loss of our model includes both classification loss and regression loss:
 \begin{equation}
\mathcal{L}_{t o t a l}=\lambda _{c l s} \mathcal{L}_{c l s}+\lambda_{reg} \mathcal{L}_{reg},
\label{eq:loss}
\end{equation}
where $\mathcal{L}_{cls}$ is the focal loss with $\gamma=2.0$ and $\alpha=0.25$. $\mathcal{L}_{reg}$ is the smooth-$l_1$ loss for TAL. 

\section{Experiments and Results}\label{sec:exp}


\subsection{Datasets and Evaluation Metrics} \label{sec:data}

\noindent\textbf{Drive\&Act}~\cite{martin2019drive} is widely used for driver activity recognition tasks. It contains 9.6 million frames across three modalities (RGB, IR, and depth) and five different camera views. The dataset provides three levels of activity labels: action units, fine-grained activities, and coarse tasks. In this paper,  we focus on the fine-grained RGB modality from the top-right view.

\noindent\textbf{SynDD2}~\cite{shaiqur2022synthetic} includes IR and RGB videos, along with annotation files, collected from three in-vehicle cameras located at the dashboard, rearview mirror, and top-right corner of the window. The dataset covers two types of activities: distracted activities and gaze zones, each with and without appearance obstructions such as hats or sunglasses.

\noindent\textbf{Evaluation Metrics.}
We follow the official evaluation metrics for Drive\&Act, using Mean-1 Accuracy (average per-class accuracy) as the primary metric, and Top-1 Accuracy for implementation assessment. For SynDD2, we evaluate temporal action localization and recognition performance using the average overlap score (AO-Score), which is defined as follows:
 \begin{equation}
o s(p, g)=\frac{\max (\min (g e, p e)-\max (g s, p s), 0)}{\max (g e, p e)-\min (g s, p s)},
\label{eq:}
\end{equation}
where $gs$ and $ge$ represent the start and end times of the ground-truth activity $g$, respectively. The variable $p$ denotes the best predicted activity of the same category as $g$, while $os$ refers to the highest overlap. The overlap between $g$ and $p$ is defined as the ratio of the intersection time to the union time of the two activities.
After matching each ground truth activity in order of their start times, any unmatched ground truth activities or unmatched predicted activities will be assigned an overlap score of 0.\looseness=-1

\begin{table}    
    \footnotesize
    \centering
     \caption{Comparison with popular methods on Drive\&Act.}
    \resizebox{.95\columnwidth}{!}{%
            \begin{tabular}{lcccccc}
                \toprule
                \multicolumn{1}{l}{\textbf{Method}} & \textbf{Modality} & \textbf{Mean-1(\%)$\uparrow$}  & \textbf{Top-1(\%)$\uparrow$}  \\ \midrule
                ResNet~\cite{he2016deep} & IR, Depth & 51.08  & 56.43  \\
                UniFormerV2~\cite{li2023uniformerv2}  & RGB, IR, Depth & 61.58 & 78.63  \\                
                MDBU (I3D)~\cite{roitberg2022comparative} & IR, NIR & 62.02 & 76.91  \\                        
                DFS~\cite{lin2024multi}  & IR, Depth & 63.12 & 77.61  \\
                \midrule
                TSM~\cite{lin2019tsm}  & IR & 59.81  & 67.75  \\
                TransDARC~\cite{pengtransdarc}  & RGB & 60.10  & 76.17  \\
                UniFormerV2~\cite{li2023uniformerv2} & RGB & 61.79 & 76.71 \\
                \cellcolor{mygray}STP (Ours) &\cellcolor{mygray}RGB &\cellcolor{mygray}\textbf{63.82} &\cellcolor{mygray}\textbf{78.32} \\
                \bottomrule
            \end{tabular}
    }
    \label{tab:drivedataset}
    \vspace{-0.2cm}
\end{table}

\begin{table}    
    \footnotesize
    \centering
     \caption{Comparison with popular methods on SynDD2.}
     \tabcolsep=0.20cm
    \resizebox{.95\columnwidth}{!}{%
            \begin{tabular}{lcccccc}
                \toprule
                \multicolumn{1}{l}{\textbf{Method}} & \textbf{Multi-View} & \textbf{Setting} & \textbf{AO-Score$\uparrow$}  \\ \midrule
                M2DAR~\cite{ma2023m2dar}  &\ding{51} & Right, Dashboard  & 0.5921  \\
                MCPRL~\cite{zhou2023multi} &\ding{51} & Right, Rear, Dashboard  & 0.6080  \\
                SKKU~~\cite{nguyen2024multi}  &\ding{51} & Right, Rear, Dashboard  & 0.7798  \\
                
                \midrule
                APC~\cite{li2023action}  &\ding{55} & Dashboard  & 0.7046  \\
                AMA~\cite{zhang2024augmented}  &\ding{55} & Right  & 0.7459  \\
                \cellcolor{mygray}STP (Ours) &\cellcolor{mygray}\ding{55}  &\cellcolor{mygray}Right  &\cellcolor{mygray}\textbf{0.7823} \\
                \bottomrule
            \end{tabular}
    }
    \label{tab:syndd2dataset}
    \vspace{-0.2cm}
\end{table}

\subsection{Implementation Details}
We utilize the pre-trained VideoMAEv2~\cite{wang2023videomae} and OpenPose~\cite{cao2017realtime} models as the backbones for video feature extraction and spatial pose estimation, respectively. Following~\cite{dong2023multi}, the input video is sampled with a temporal stride of $8$, each frame is resized to $224\times224$, and only 13 key points are used per frame. The Multi-Scale Encoder consists of $6$ layers,  with $4$ heads and $256$-dimensional embeddings. In the training stage, we use AdamW~\cite{loshchilov2017decoupled} optimizer with an initial learning rate of 1e-3 and cosine decay learning rate strategy~\cite{loshchilov2016sgdr} with power set to $0.9$. 
During inference, the initial predictions are screened by SoftNMS~\cite{bodla2017soft} with a threshold of 0.2.
The weights $\lambda _{cls}$ and $\lambda _{reg}$ are set as 1 and 1.5, respectively.

\begin{table}
        \caption{Effects of each component in our method.}
        \tabcolsep=0.15cm
	\resizebox{.95\columnwidth}{!}{
	\begin{tabular}{ccccccccc}
		\toprule
		 \textbf{Spatial}  & \textbf{Temporal} & \textbf{Spatial-Temporal}   & \textbf{Causal-aware}  & \multirow{2}{*}{\textbf{AO-Score$\uparrow$}}  \\
            \textbf{Feature} & \textbf{Feature} &  \textbf{Feature} & \textbf{Moudel} &  \\
		\midrule
		 \ding{51}              &                  &   &  & 0.7223  \\
                       & \ding{51}             &  &  & 0.7298   \\
		            &     & \ding{51}& & 0.7443   \\
                     &   & \ding{51}& \ding{51} & \textbf{0.7823}   \\
		\bottomrule
	\end{tabular}}
        \label{tab:overall-ablation}
        \vspace{-0.2cm}
\end{table}

\begin{table}    
    \centering
     \caption{The comparison of the model efficiency results.}
     \tabcolsep=0.20cm
    \resizebox{.95\columnwidth}{!}{%
            \begin{tabular}{c|l|ccccc}
                \toprule
                \multicolumn{1}{c|}{\textbf{Modality}} & \textbf{Methods} & \textbf{Latency(ms)$\downarrow$}  & \textbf{\#Param$\downarrow$}  \\ 
                \hline\hline
                Dual&UniFormerV2~\cite{li2023uniformerv2}   & 33.0 & 47.2M  \\                                       
                Dual&DFS~\cite{lin2024multi}   & 28.0 & 38.8M  \\
               
                Single&TSM~\cite{lin2019tsm}   & 15.0  & 25.3M  \\
                Single&I3D~\cite{roitberg2022comparative}   & 18.3  & 28.0M  \\
                Single&STP (Ours)  &\textbf{14.2} &\textbf{23.7M} \\
                \bottomrule
            \end{tabular}
    }
    \label{tab:efficiency}
    \vspace{-0.2cm}
\end{table}

\subsection{Quantitative Results}
Table~\ref{tab:drivedataset} shows the results of our method on the Drive\&Act test dataset. We compare popular single-modal and multi-modal driving action recognition methods. 
STP significantly outperforms all methods, achieving a Mean-1 accuracy of 63.82\% and a Top-1 accuracy of 78.32\%.
This shows that our method can effectively achieve high-precision driver action recognition even without additional information input.

Table~\ref{tab:syndd2dataset} shows the results on the SynDD2 dataset. In this table, we categorize and compare the methods using different camera angles. 
Without the need for complex multi-view fusion, our approach achieved a state-of-the-art performance with 0.7823 AO-Score, demonstrating the effectiveness of our proposed Spatial-Temporal Perception. It is worth noting that our method only relies on the RGB input of the camera on the right side of the driver, which greatly reduces the hardware cost of the actual scene.

\subsection{Ablation Study and Visualization}
To validate the effectiveness of our STP model, we conducted several ablation experiments on the SynDD2 dataset. 

\noindent\textbf{Ablation Study.}
We present the results of the ablation studies in Table~\ref{tab:overall-ablation}. We progressively add the spatial-temporal perception structure and the causal-aware module, and report the corresponding AO-Score. The results indicate that using spatial or temporal features alone provides only limited improvements. In contrast, the proposed spatial-temporal perception significantly enhances the AO-Score. Furthermore, the causal-aware module effectively captures relationships between video frames, leading to additional performance gains.

\noindent\textbf{Efficiency comparison.}
Model efficiency is critical for real-time driver monitoring systems. We further evaluate the efficiency of the model in terms of latency and parameter size, as shown in Table~\ref{tab:efficiency}. For a fair comparison, we categorize the current popular methods into dual and single modality inputs, ensuring consistent input cropping. The results demonstrate that our method not only achieves superior performance but also retains the efficiency benefits of lower latency and a reduced parameter count typical of single-modality inputs.

\noindent\textbf{Results Visualization.}
As shown in Table~\ref{tab:visualization},we further visualize several challenging cases and their corresponding results during the keypoint detection stage. These examples clearly demonstrate that changes in the distance between key points (such as between the fingers and mouth) provide valuable prior knowledge, enabling the model to make accurate inferences. Even when actions appear similar, our method can accurately distinguish between the driver's Calling and Eating actions and predict the precise start and end times of these actions.

\begin{table}    
   \large
    \centering
     \caption{Sample visualizations of the process of keypoint detection.}
     \tabcolsep=0.15cm
    \resizebox{.98\columnwidth}{!}{%
            \begin{tabular}{m{12cm}|m{1.5cm}}
                \toprule
                \textbf{Input}  & \textbf{Results}  \\ 
                \hline\hline
        
                 \includegraphics[width=12cm]{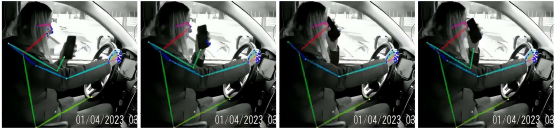} & \makecell[c]{Calling}  \\
                \midrule
                 \includegraphics[width=12cm]{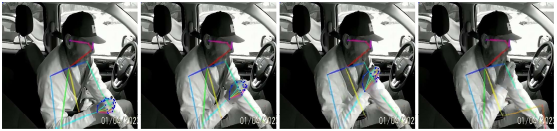} & \makecell[c]{Eating} \\
                  \midrule
                 \includegraphics[width=12cm]{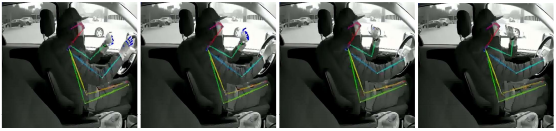} & \makecell[c]{Drinking} \\ 
                \bottomrule
            \end{tabular}
    }
    \label{tab:visualization}
    \vspace{-0.2cm}
\end{table}

\section{Conclusion}\label{sec:conc}
In this paper, we introduce a novel Spatial-Temporal Perception (STP) architecture designed to enhance action recognition and temporal action localization by capturing both the temporal dynamics and spatial relationships between key objects. Unlike multimodal approaches, STP directly extracts temporal and spatial distance features from RGB video clips, encoding these dual features by optimizing the likelihood across various factorization orders. This integration allows STP to detect subtle behavioral changes, even in complex scenarios. Furthermore, the inclusion of a causal-aware module improves detection efficiency by exploring the relationships between video frame features. Validated on two publicly available driver distraction detection benchmarks, our approach achieves state-of-the-art performance, highlighting its effectiveness and potential for broader applications.

\section*{Acknowledgment}
This work was supported by Guangzhou-HKUST(GZ) Joint Funding Program(Grant No.2023A03J0008), Education Bureau of Guangzhou Municipality.
\newpage
{\small
\bibliographystyle{ieee}
\bibliography{egbib}
}

\end{document}